# Multi-models with averaging in feature domain for non-invasive blood glucose estimation

Yiting Wei
*Faculty of Information Engineering*
*Guangdong University of technology*
*Guangzhou, 510006, China.*
1464684152@qq.com

Bingo Wing-Kuen Ling
*Faculty of Information Engineering*
*Guangdong University of technology*
*Guangzhou, 510006, China.*
yongquanling@gdut.edu.cn

Qing Liu
*Faculty of Information Engineering*
*Guangdong University of technology*
*Guangzhou, 510006, China.*
Liuq@gdut.edu.cn

Jiaxin Liu
*Faculty of Information Engineering*
*Guangdong University of technology*
*Guangzhou, 510006, China.*
726310129@qq.com

*Abstract*—Diabetes is a serious chronic metabolic disease. In the recent years, more and more studies focus on the use of the non-invasive methods to achieve the blood glucose estimation. More and more consumer technology enterprises focusing on human health are committed to implementing accurate and non-invasive blood glucose algorithm in their products. The near infrared spectroscopy built in the wearable devices is one of the common approaches to achieve the non-invasive blood glucose estimation. However, due to the interference from the external environment, these wearable non-invasive methods yield the low estimation accuracy. Even if it is not medical equipment, as a consumer product, the detection accuracy will also be an important indicator for consumers. To address this issue, this paper employs different models based on different ranges of the blood glucose values for performing the blood glucose estimation. First the photoplethysmograms (PPGs) are acquired and they are denoised via the bit plane singular spectrum analysis (SSA) method. Second, the features are extracted. For the data in the training set, first the features are averaged across the measurements in the feature domain via the optimization approach. Second, the random forest is employed to sort the importance of each feature. Third, the training set is divided into three subsets according to the reference blood glucose values. Fourth, the feature vectors and the corresponding blood glucose values in the same group are employed to build an individual model. Fifth, for each feature, the average of the feature values for all the measurements in the same subset is computed. For the data in the test set, first, the sum of the weighted distances between the test feature values and the average values obtained in the above is computed for each model. Here, the weights are defined based on the importance sorted by the random forest obtained in the above. The model corresponding to the smallest sum is assigned. Finally, the blood glucose value is estimated based on the corresponding model. Compared to the state of arts methods, our proposed method can effectively improve the estimation accuracy. In particular, the mean absolute relative difference (MARD) and the percentage of the data fall in the zone A of the Clarke error grid yielded by our proposed method reaches 12.19%, and 87.0588%, respectively.

## I. Introduction

According to the 10th edition of the World Diabetes Map released by the International Diabetes Federation (IDF), in 2021, 537 million (10.5%) adults aged 20-79 will have Type 1 diabetes [1]. The total number of people with diabetes is expected to increase to 643 million (11.3%) by 2030 and 783 million (12.2%) by 2045 [1]. Diabetes is a chronic metabolic disease characterized by hyperglycemia, which can cause vascular and neural lesions, resulting in various complications and seriously threatening the health and life of patients [2].

Diabetic patients must monitor their blood glucose regularly every day. The most traditional monitoring method is to collect fingertip blood with invasive methods, but it will bring great burden to the patients' body and mind [3]. Gradually, the concepts of micro-invasive blood glucose monitoring and continuous blood glucose monitoring have been proposed, but these methods have disadvantages such as physiological delay and inapplicability to patients with immune diseases [4]. Therefore, non-invasive blood glucose monitoring has gradually become a new focus of attention.

With regard to the research status of non-invasive blood glucose estimation, some scholars have proposed the following methods. First, using the bioimpedance spectroscopy method, according to the principle of non-ionic soluble substances of glucose, electromagnetic wave radiation is released to glucose, and the characteristic value of electromagnetic wave of the absorbed frequency is extracted. The glucose concentration in blood can be determined through quantitative analysis [8]. However, this method does not achieve a real sense of non-invasive, and the accuracy is low. At the same time, from the perspective of consumer electronics product design, this method requires regular replacement of the sensor probe, which will have concerns about convenience and cost for consumers. In addition, there are ways to measure blood glucose through blood substitutes such as body fluids. Among them, measuring the glucose concentration in tears is a representative method [9]. However, some studies have shown that the correlation between blood glucose concentration and the glucose concentration in tears is weak. At the same time, the instability of tear formation and evaporation will affect the accuracy of measurement. Metabolic heat conformation method has also been applied to non-invasive blood glucose estimation. It uses multi-sensor integration technologies to digitize the temperature, humidity, blood flow rate and blood oxygen saturation that reflect metabolic heat, and then calculates the functional relationship between metabolic heat, blood glucose concentration and oxygen supply to finally obtain the blood glucose value [10]. However, this method requires many parameters to be measured, which makes sensor integration difficult and parameter coupling difficult. At the same time, it is very vulnerable to environmental temperature and other conditions, which will affect the accuracy of the results. In addition, there is a method based on the reverse iontophoresis method to measure the glucose concentration in the subcutaneous tissue fluid [11], but it will cause the discomfort of electric current burning the skin. At the same time, the glucose concentration in the tissue fluid will slightly lag behind the concentration in the blood, which will cause low measurement accuracy.

At present, optical methods are the most mainstream in non-invasive blood glucose research, among which near-infrared spectroscopy is the most promising [5]. The non-invasive detection of blood glucose concentration by near-infrared spectroscopy is based on Lambert Beer's law and relies on the specificity of glucose light absorption to measure blood glucose concentration. According to the hydrogen bond of glucose in the near-infrared spectral region (the wavelength range is about 780nm~2526nm), such as the vibration combination and doubling of C-H, N-H, O-H, we can obtain rich near-infrared spectral information of substances[12]. PPG (Photoplethysmographic) is based on near-infrared spectroscopy and LED light source and detector to measure the attenuated light reflected and absorbed by human blood vessels and tissues, record the pulsation state of blood vessels and measure pulse wave[13].Research shows that PPG signal can reflect rich blood glucose information[14]. In order to better apply these principles to practice, it is the most direct way to obtain PPG signals of the human body through wearable devices equipped with optical sensors. More and more consumer technology enterprises focusing on human health are committed to developing high-precision non-invasive blood glucose detection algorithms that meet the standards. However, due to external environmental factors, wearing methods and individual differences [6], the accuracy of non-invasive blood glucose devices is still insufficient. However, for patients with diabetes and potential users, the accuracy of non-invasive blood glucose detection products is always the most concerned part, because it is related to the life and health of patients. In addition, in the process of product launch and promotion, the application for certification of medical devices also has strict requirements on the accuracy of products. Based on this, this paper aims to propose a non-invasive blood glucose estimation method based on averaging in the feature domain and piecewise domain classification modeling, and is committed to providing a new idea of accurate algorithm design for the industry.

This paper proposes a multi-models with averaging in feature domain for non invasive blood glucose estimation method. After averaging the feature domain data and providing a multi-models fusion modeling method, the computer numerical simulation results show that the method proposed in this paper can effectively improve the accuracy of blood glucose estimation. The outline of the paper is as follows. Section II presents our proposed method. Section III presents the computer numerical simulation results. Finally, the conclusion is drawn in Section IV.

## II. OUR PROPOSED METHODS

### A. Data acquisition

First of all, for signal acquisition, we use three groups of optical sensors with wavelengths of 890nm, 1450nm and 1650nm to collect PPGs, and their sampling rates are 1000Hz, 50Hz and 50Hz respectively. For the reference blood glucose value, we use the BeneCheck blood glucose meter certified by the Food and Drug Administration (FDA).

Our experimental process is shown in Figure 1. A total of eight volunteers were recruited to collect data for twelve days four times each day. In order to further expand the scope of data and verify the effectiveness of the experimental design, we arranged volunteers to conduct a blood glucose comparison experiment on different eating behaviors during the experimental period, including drinking cola after meals, eating normal meals and eating ketogenic foods. We finally collected a total of 465 groups of experimental data.

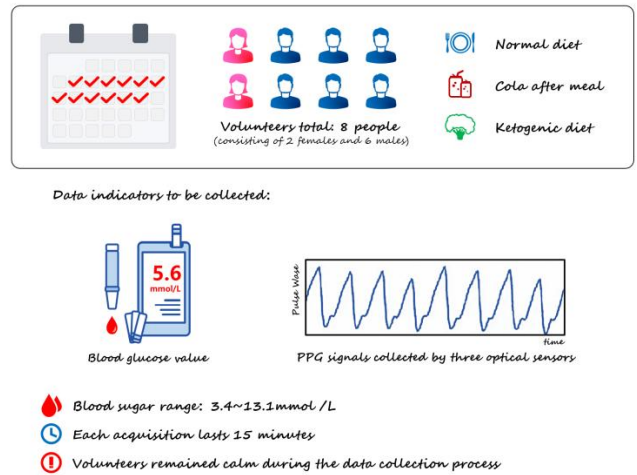

Fig. 1. The intelligent watch for acquiring the PPGs.

### B. Denoising

We use bit plane SSA method to denoise the collected PPG signals, and the denoising results are shown in Figure 2.

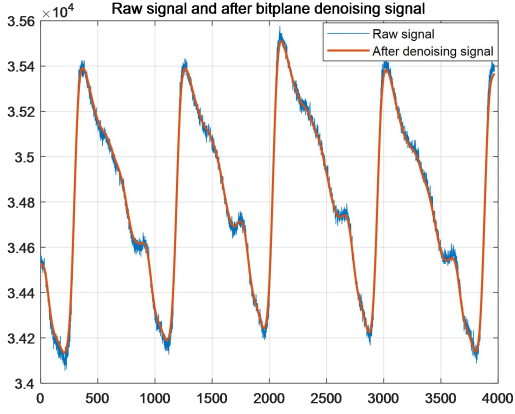

Fig. 2. Effect comparison before and after signal denoising.

*C. Feature extraction*

Our features mainly come from the three PPG signals collected. After detecting the feature points of PPG signals, we have extracted 24 time-domain features and 7 frequency-domain features. Figure 3 shows the feature points of some PPG signals we extracted.

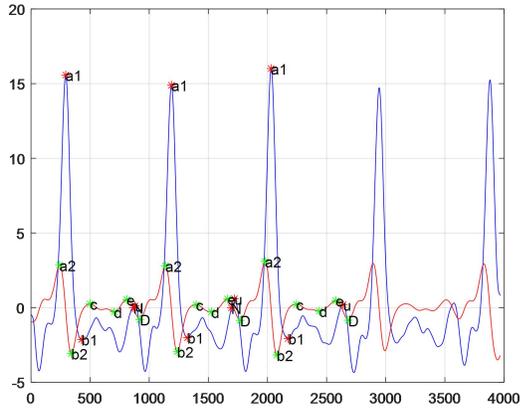

Fig. 3. Part of feature points of PPG signals extracted.

*D. Averaging in the feature domain*

- Step0: Initialize the feature index $k=1$.
- Step1: Call the original feature matrix $X$.

$$X = \begin{bmatrix} x_{1,1} & x_{1,2} & \cdots & x_{1,K} \\ x_{2,1} & x_{2,2} & \cdots & x_{2,K} \\ \vdots & \vdots & & \vdots \\ x_{N,1} & x_{N,2} & \cdots & x_{N,K} \end{bmatrix}. \quad (1)$$

The dimensions of the original feature matrix are $N \times K$, that is, there are $K$ features and $N$ groups of measurements. Denote the $k^{th}$ feature as $x_k, k \in [1, K]$. Find the weight matrix $W_k$, which satisfies the equation

$$y_k = W_k x_k, \quad (2)$$

$y_k$ represents the $k^{th}$ feature after performing the averaging.

- Step2: Let $g$ be the vector of the reference blood glucose values. Solving the optimization problems

$$\max_{W_k} \frac{x_k^T W_k^T g g^T W_k x_k}{g^T g x_k^T W_k^T W_k x_k}, \quad (3)$$

subject to

$$g^T g x_k^T W_k^T W_k x_k = C_k. \quad (4)$$

where $C_k$ is guaranteed to be a constant.

- Step3: Let $L$ be the window length, let $w_k$ be the vector of the weight coefficients, and let

$$X_k = \begin{bmatrix} x_{1,k} & x_{2,k} & \cdots & x_{L,k} \\ x_{2,k} & x_{3,k} & \cdots & x_{L+1,k} \\ \vdots & \vdots & \ddots & \vdots \\ x_{N-L+1,k} & x_{N-L+2,k} & \cdots & x_{N,k} \end{bmatrix}. \quad (5)$$

Then, the optimization problem defined in (3) becomes the following optimization problem:

$$\max_{w_k} w_k^T X_k^T g g^T X_k w_k, \quad (6)$$

subject to

$$w_k^T X_k^T X_k w_k = \frac{C_k}{g^T g}. \quad (7)$$

Let

$$S_{A,k} = X_k^T g g^T X_k, \quad (8)$$

$$S_{B,k} = X_k^T X_k, \quad (9)$$

$$\tilde{C}_k = \frac{C_k}{g^T g}. \quad (10)$$

Assume

$$rank(S_{B,k}) = L, \quad (11)$$

then,

$$\max_{w_k} J(w_k) = w_k^T S_{A,k} w_k - \lambda_k (w_k^T S_{B,k} w_k - \tilde{C}_k), \quad (12)$$

$$\frac{\partial J}{\partial w_k} = 2 S_{A,k} w_k - 2\lambda_k S_{B,k} w_k = 0, \quad (13)$$

then we have,

$$S_{B,k}^{-1} S_{A,k} w_k = \lambda_k w_k. \quad (14)$$

- Step4: Let

$$Q_k = S_{B,k}^{-1} S_{A,k}, \quad (15)$$

then let

$$u_k = X_k^T g, \quad (16)$$

$$v_k = S_{B,k}^{-1} u_k, \quad (17)$$

$$Q_k = v_k u_k^T, \quad (18)$$

$$u_k v_k^T w_k = \lambda_k w_k, \quad (19)$$

in which

$$u_k = w_k, \quad (20)$$

$$v_k^T w_k = \lambda_k. \quad (21)$$

- Step5: Substitute $w_k$ into (2) to obtain $y_k$.
- Step6: Increment the value of $k$. Go to step1 until $k = K$.

### III. COMPUTER NUMERICAL SIMULATION RESULTS

#### A. Piecewise domain modeling method based on random forest

- After averaging the $k^{th}$ feature in the training set, let $\tilde{y}_k$ be the vector containing the $k^{th}$ feature values for all the measurements in the training set. Let $\tilde{y}_{j,k}$ be the $k^{th}$ feature value at the $j^{th}$ sorted measurement. That is,

$$\tilde{y}_k = \begin{bmatrix} \tilde{y}_{1,k} & \cdots & \tilde{y}_{N,k} \end{bmatrix}^T. \quad (22)$$

Then, $\tilde{y}_k$ is normalized to the unit energy vector. Let

$$q_k = \frac{1}{\sqrt{\tilde{y}_k^T \tilde{y}_k}} \quad (23)$$

be the normalized gain. Let $\hat{y}_k$ be the normalized vector. Then, we have

$$\hat{y}_k = q_k \tilde{y}_k. \quad (24)$$

On the other hand, for each test vector in the test set, the $k^{th}$ feature value is multiplied by $q_k$ to obtain the re-scaled feature value.

- All the data were randomly divided into 3:1, of which 75% is training set and 25% is test set.

- Then, random forest regression algorithm is used for the first prediction to obtain the importance ranking and contribution rate of each feature. Then, reorder the feature4s and generate three sub-training sets according to the blood glucose value namely $RF_1$, $RF_2$ and $RF_3$ from top to bottom. After that, derive the corresponding weight value according to the contribution rate of each feature. The dimensions of the test set are $M \times K + 1$. Let the $m^{th}$ group of eigenvector elements in the test set are $\tilde{y}_{m,1}, \tilde{y}_{m,2} \ldots \ldots \tilde{y}_{m,k}$, the corresponding feature weights are $a_1, a_2 \ldots \ldots a_k$, the corresponding blood glucose value are $BG_1, BG_2 \ldots \ldots BG_k$. Next, calculate the average value of each feature of $RF_1$, denoted as $\bar{y}_{1,1}, \bar{y}_{1,2} \ldots \ldots \bar{y}_{1,k}$. Similarly, calculate the average value of each feature of $RF_2$ and $RF_3$, denoted as $\bar{y}_{2,1}, \bar{y}_{2,2} \ldots \ldots \bar{y}_{2,k}$ and $\bar{y}_{3,1}, \bar{y}_{3,2} \ldots \ldots \bar{y}_{3,k}$ respectively. Then, calculate three groups of Euclidean distances $x_{m,1}, x_{m,2}, x_{m,3}$ for the $m^{th}$ measurement of the test set.

$$x_{m,1} = \sqrt{a_1(\tilde{y}_{m,1} - \bar{y}_{1,1})^2 + a_2(\tilde{y}_{m,2} - \bar{y}_{1,2})^2 + \ldots + a_k(\tilde{y}_{m,k} - \bar{y}_{1,k})^2}, \quad (25)$$

$$x_{m,2} = \sqrt{a_1(\tilde{y}_{m,1} - \bar{y}_{2,1})^2 + a_2(\tilde{y}_{m,2} - \bar{y}_{2,2})^2 + \ldots + a_k(\tilde{y}_{m,k} - \bar{y}_{2,k})^2}, \quad (26)$$

$$x_{m,3} = \sqrt{a_1(\tilde{y}_{m,1}-\overline{y}_{3,1})^2 + a_2(\tilde{y}_{m,2}-\overline{y}_{3,2})^2 + \ldots + a_k(\tilde{y}_{m,k}-\overline{y}_{3,k})^2} \ . \quad (27)$$

Compare the sizes of $x_{m,1}, x_{m,2}$ and $x_{m,3}$, if $x_{m,t}$ is the smallest, then the $m^{th}$ measurement of the test set will be classified to Class $t$, $t \in [1,3]$. Here, the classification of the test set has been completed. The specific algorithm flow is shown in the Figure 4.

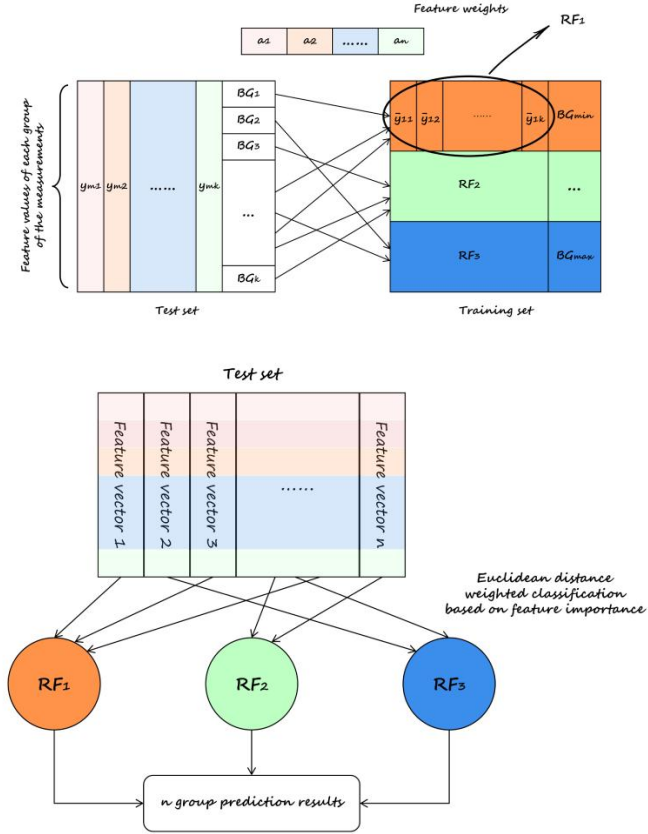

Fig. 4. The piecewise domain modeling algorithm flow.

- The divided test set and three training sets are one-to-one corresponded to establish three random forest regression models to obtain three groups of blood glucose prediction results. The results are spliced to obtain complete blood glucose prediction results.

### B. Computer numerical simulations

In order to compare the results, the raw data without performing the averaging are randomly divided into 3:1, and a group of blood glucose prediction results are obtained by only using the random forest. We present the results without any processing and the results obtained from the joint averaging in the feature domain and the piecewise feature selection method in Table I. It is obvious from various indicators that the method proposed in this paper can effectively improve the effect of blood glucose estimation. At the same time, we use Clarke error grid [7] to draw the results based on the method proposed in this paper, as shown in Figure 5, 87.0588% of the data fall in zone A of the grid.

TABLE I. RESULTS OF VARIOUS INDICATORS OF THE TWO METHODS

| Methods | Indicators | | | |
|---|---|---|---|---|
| | $R$ | $MAE \pm SD$ (mmol/L) | $RMSE$ (mmol/L) | $MARD$ (%) |
| Random forest regression | 0.4911 | $1.2361 \pm 1.9801$ | 1.9677 | 16.71 |
| Our proposed method | 0.7685 | $0.8484 \pm 1.5838$ | 1.7667 | 12.19 |

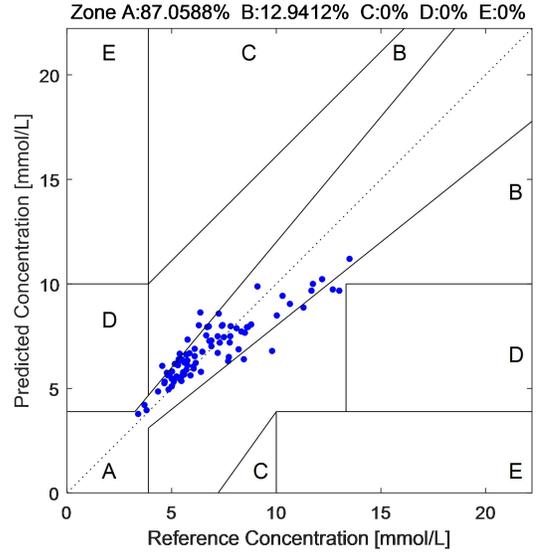

Fig. 5. Clarke error grid of non-invasive blood glucose estimation method based on the joint averaging in the feature domain and the piecewise feature selection method.

### IV. CONCLUSION

Non-invasive blood glucose estimation is a hot topic for the huge volume of diabetic patients. The study of non-invasive blood glucose detection algorithm also has important practical significance and value. This paper mainly proposes a joint averaging in the feature domain and the piecewise feature selection method for performing the blood glucose estimation. It can eliminate the data error caused by improper environment and operation in the feature domain. After that, complete the data classification task according to the importance of features, and then use the multi-models method to predict the blood glucose value. The pretreatment of feature domain data and the improvement of regression modeling method can effectively improve the accuracy of blood glucose estimation. According to the results of computer numerical simulation, the method proposed in this paper has significantly improved the prediction results compared with the simple random forest regression model, 87.0588% of the data fell on zone A of Clarke Error Grid, the Pearson correlation coefficient R reached 0.7685, and the Mean Absolute Relative Difference (MARD) reached 12.19%. Next, further experiments can be conducted from two aspects: further expanding the data set and using a variety of machine learning models to verify, then the experimental design can be further optimized.

ACKNOWLEDGMENT

This paper was supported partly by the National Nature Science Foundation of China (no. U1701266, no. 61671163, no. 62071128 and no. 61901123), the Team Project of the Education Ministry of the Guangdong Province (no. 2017KCXTD011), the Guangdong Higher Education Engineering Technology Research Center for Big Data on Manufacturing Knowledge Patent (no. 501130144) and the Hong Kong Innovation and Technology Commission, Enterprise Support Scheme (no. S/E/070/17).